\documentclass[conferencel]{IEEEtran}
\usepackage{cite}
\usepackage{amsmath,amssymb,amsfonts}
\usepackage{graphicx}
\usepackage{textcomp}
\usepackage{hyperref}
\usepackage{mathtools}
\usepackage{booktabs}
\usepackage{multirow}
\usepackage{pifont}
\usepackage{graphbox}
\usepackage{algorithm,algorithmicx,algpseudocode}

\def\BibTeX{{\rm B\kern-.05em{\sc i\kern-.025em b}\kern-.08em
    T\kern-.1667em\lower.7ex\hbox{E}\kern-.125emX}}

\begin{document}
\title{Latent Diffusion for Medical Image Segmentation: End to end learning for fast sampling and accuracy}
\author{Fahim Ahmed Zaman, Mathews Jacob, Amanda Chang, Kan Liu, Milan Sonka and Xiaodong Wu
\thanks{This research was supported in part by NIH Grants R01HL171624, R01AG067078 and R01EB019961}
\thanks{Fahim Ahmed Zaman, Milan Sonka and Xiaodong Wu are with the department of Electrical and Computer Engineering, University of Iowa, IA 52240, USA (email: \{fahim-zaman, milan-sonka, xiaodong-wu\}@uiowa.edu.)}
\thanks{Mathews Jacob is with the department of Electrical and Computer Engineering, University of Virgina, VA 22904, USA (email: mjacob@virginia.edu.)}
\thanks{Amanda Chang is with the department of Medicine, University of Iowa, IA 52240, USA (email: amanda-chang@uiowa.edu.)}
\thanks{Kan Liu is with the Washington University School of Medicine, MO 63110, USA (email: kanl@wustl.edu.)}}

\maketitle

\begin{abstract}
Diffusion Probabilistic Models (DPMs) suffer from inefficient inference due to their slow sampling and high memory consumption, which limits their applicability to various medical imaging applications. In this work, we propose a novel conditional diffusion modeling framework (LDSeg) for medical image segmentation, utilizing the learned inherent low-dimensional latent shape manifolds of the target objects and the embeddings of the source image with an end-to-end framework. Conditional diffusion in latent space not only ensures accurate image segmentation for multiple interacting objects, but also tackles the fundamental issues of traditional DPM-based segmentation methods: (1) high memory consumption, (2) time-consuming sampling process, and (3) unnatural noise injection in the forward and reverse processes. 
The end-to-end training strategy enables robust representation learning in the latent space related to segmentation features, ensuring significantly faster sampling from the posterior distribution for segmentation generation in the inference phase.
Our experiments demonstrate that LDSeg achieved state-of-the-art segmentation accuracy on three medical image datasets with different imaging modalities.
In addition, we showed that our proposed model was significantly more robust to noise compared to traditional deterministic segmentation models. The code is available at \href{https://github.com/FahimZaman/LDSeg.git}{https://github.com/FahimZaman/LDSeg.git}.
\end{abstract}


\section{Introduction}
\label{introduction}
In the field of medical imaging, image segmentation is a crucial step in identifying and monitoring disease-related pathologies, clinical decisions in treatment, and the evaluation of disease progression \cite{Preeti}. Traditional deep learning (DL) based segmentation models have achieved impressive accuracy in various imaging modalities, which often match/outperform field level experts \cite{Hesamian, wang2022}. These DL-based models, mostly including convolutional neural networks (CNN), and vision transformers (ViT) are generally trained end-to-end in a discriminative manner. Recently, generative models have emerged as powerful image segmentation tools, taking advantage of learning the underlying statistics of target objects, conditioned on the source image. These conditional generative models include diffusion probabilistic models (DPMs)~\cite{ho2020denoising, ho2022cascaded, song2019generative, song2021scorebased} and generative adversarial networks (GAN). 

In computer vision, DPMs have achieved remarkable results for image generation, outperforming other generative models \cite{dhariwal2021diffusion}. Standard DPMs have two major steps: a forward process that perturbs the image with added Gaussian noise and a reverse process that starts with a Gaussian noise and iteratively denoises the image to generate a clean image of the original data distribution. The learned denoiser is trained with noisy images for different noise variances, thus modeling the gradient of the smoothed log prior (score) of the images. Recently, intensive efforts have been made to extend DPMs for medical image segmentation \cite{wu2022medsegdiff,wu2023medsegdiff,bogensperger2023scorebased,Rahman2023AmbiguousMI, Ding_2024, chen2024hidiffhybriddiffusionframework}. The DPMs used for segmentation differ from the ones used for image generation in the sense that the forward and reverse processes involve the noise addition and denoising of the segmentation mask instead of the source image. In particular, the denoising network models the gradient of the smoothed conditional distribution of the segmentation labels given the source image. The objective of the segmentation DPM is thus to sample a segmentation mask from the conditional distribution.

To fully harness the power of DPMs for medical image segmentation, some fundamental challenges must be addressed. First, the original DPM formulation assumes that the pixel values are continuous variables, compared to binary/integer segmentation labels in the segmentation setting. The above denoising strategy using Gaussian noise may not learn the true score in our setting. This mismatch (continuous values versus semantic labels) results in discrepancies during inference. For example, additional thresholding is needed to obtain the final segmentation mask by filling in hole-shaped structures that are caused by high-frequency noises~\cite{bogensperger2023scorebased}. Wu {\em et al.} proposed the use of frequency parser blocks in the hidden layers of the denoiser to modulate high-frequency noise \cite{wu2022medsegdiff}, but it does not guarantee clean results after sampling and may need post-processing. Bogensperger {\em et al.} proposed to transform the discrete segmentation mask into a signed distance function (SDF-DDPM), where each pixel represents the signed Euclidean distance from the boundary of the closest object \cite{bogensperger2023scorebased}. A limitation of this approach is that the distance map for the multi-class images is ambiguous. Zaman {\em et al.} proposed to re-parameterize the segmentation masks with a graph structure which guarantees naturally continuous perturbations for surface positions on the graph columns \cite{zaman2023surf}. However, this model suffers from the multiclass mask representation problem, as it poses great challenges to have a common graph column structure to define surface positions for different objects. There is a compelling need to have a proper reparameterization technique that can be implemented for multiclass objects and guarantees smooth state transitions between classes. 

The second key challenge of DPMs is the time-consuming iterative sampling process, which makes segmentation DPMs significantly slower than their deterministic counterparts. Various approaches have been developed to improve the sampling efficiency for natural image generation \cite{nichol2021improved, song2020denoising, lu2022dpm, wang2023learning}. Recently, latent diffusion models, which perform the sampling in a low-dimensional latent space, have been used to speed up the diffusion process in natural image generation~\cite{rombach2022high} and segmentation~\cite{ldznet}. In both works, the latent representation of the source image was used. Liu  {\em et al.} proposed a latent diffusion model that uses label rectification in latent space for semi-supervised medical image segmentation \cite{Liu_DiffRect_MICCAI2024}. Latent diffusion has also been used in audiovisual segmentation for robust audio recognition~\cite{mao2023contrastive}. Recently, Vu Quoc {\em et al.} proposed a latent diffusion model for image segmentation \cite{vu2023lsegdiff}. They proposed a two-step training strategy in which a variational-autoencoder (VAE) is first trained to learn the latent distribution of the label image, followed by training a conditional denoiser of the latent codes; the embedding of the source image is used as the condition. This latent diffusion model is computationally efficient in comparison to traditional DPMs. However, since the score / denoiser model is trained separately by minimizing the mean square error (MSE) between the recovered and true latents, the MSE loss in the latent domain may not well capture the segmentation errors, which may compromise the accuracy of segmentation. Vahdat {\em et al.} proposed to use an end-to-end training strategy by jointly learning latent embeddings and a denoiser of latent codes for image generation \cite{vahdat2021score}. They observed improved accuracy and faster sampling for the end-to-end framework.

We introduce a novel conditional latent diffusion-based generative framework (LDSeg) for medical image segmentation, which capitalizes on the inherent advantages of latent diffusion models. Unlike the two-step training strategy in \cite{vu2023lsegdiff}, we jointly train the encoder, decoder, and the score model in an end-to-end fashion. We use a combination of denoiser loss in the latent domain and segmentation loss in the label domain. The proposed LDSeg learns the standardized latent representation of the target object shape manifolds, enabling smooth state transitions between object classes. Moreover, unlike traditional DPMs, LDSeg learns to sample from the posterior distribution, which is significantly simpler and more concentrated than the prior distribution. Clearly, sampling from a narrow/concentrated distribution would be faster.
The major contributions of this work are summarized as follows.
\begin{itemize}
  \item To the best of our knowledge, this is the first work to leverage the jointly learned latent representation of the image and target object's shape manifolds with a diffusion denoiser in latent space, for robust posterior prediction. Our experiments show that the end-to-end training strategy offers improved performance and faster sampling, compared to the two-step training scheme.
  \item The label encoder maps the continuous domain latent variables to a standardized latent space ($\mu=0, \sigma=1$). Hence, standard diffusion theory can be applied to the latents, unlike in the case of discrete labels. For instance, this mitigates the need for post-processing of the estimated labels to remove hole-shaped structures. In addition, the standardized distribution of the latents (having a similar distribution of the added Gaussian noise) will ensure faster convergence and improved gradient flow during training.
  \item The diffusion in the latent space ensures less memory consumption and much faster training and inference process, enabling efficient application of DPM on 3- and higher-dimensional medical image segmentation.
  \item The LDSeg model is significantly more robust to noise in the source images compared to the deterministic segmentation models due to the low-dimensional image embeddings, which mitigates the segmentation challenges in medical images with noisy acquisition.
\end{itemize}

\section{Background}
\subsection{Denoising Diffusion Probabilistic Models (DDPM)}
DDPMs are designed to learn a data distribution by gradually denoising a normally distributed variable, which corresponds to learning the reverse process of a fixed-length Markov chain $T$. More precisely, this denoising reverse process can be modeled as $p_{\theta}(x_{0:T})$, which is a Markov chain with learned Gaussian transitions starting at $p(x_T) \sim \mathcal{N}(x_T;\mathrm{0,I})$:
\begin{equation}
    p_{\theta}(x_{0:T}) \coloneqq p(x_T) \prod_{t=1}^{T} p_{\theta}(x_{t-1} \mid x_t),
\end{equation}
\begin{equation}
    p_{\theta}(x_{t-1} \mid x_t) \coloneqq \mathcal{N}(x_{t-1}; \mu_{\theta}(x_t,t), \Sigma_{\theta}(x_t,t))
\end{equation}
where $x_0 \sim q(x_0)$ is a sample from a real data distribution, $x_1, \dotsc ,x_T$ are transitional states for timesteps $t= 1, \dotsc ,T$.

The forward process in a DDPM is also a Markov chain, which gradually adds noise to the image. Given data $x_0 \sim q(x_0)$ sampled from the real distribution, the forward process at time $t \in [1, T ]$ can be defined as $q(x_t \mid x_{t-1})$, where Gaussian noise is gradually added given a noise variance schedule $\beta_t \in [\beta_1, \beta_T]$:
\begin{equation}
    q(x_{1:T} \mid x_0) \coloneqq \prod_{t=1}^{T} q(x_{t} \mid x_{t-1}),
\end{equation}

\begin{equation}
    q(x_{t} \mid x_{t-1}) = \mathcal{N}(x_t; \sqrt{1-\beta_t}x_{t-1}, \beta_{t} I)
\end{equation}

\begin{figure*}[ht]
    \centering
    \includegraphics[width=7in]{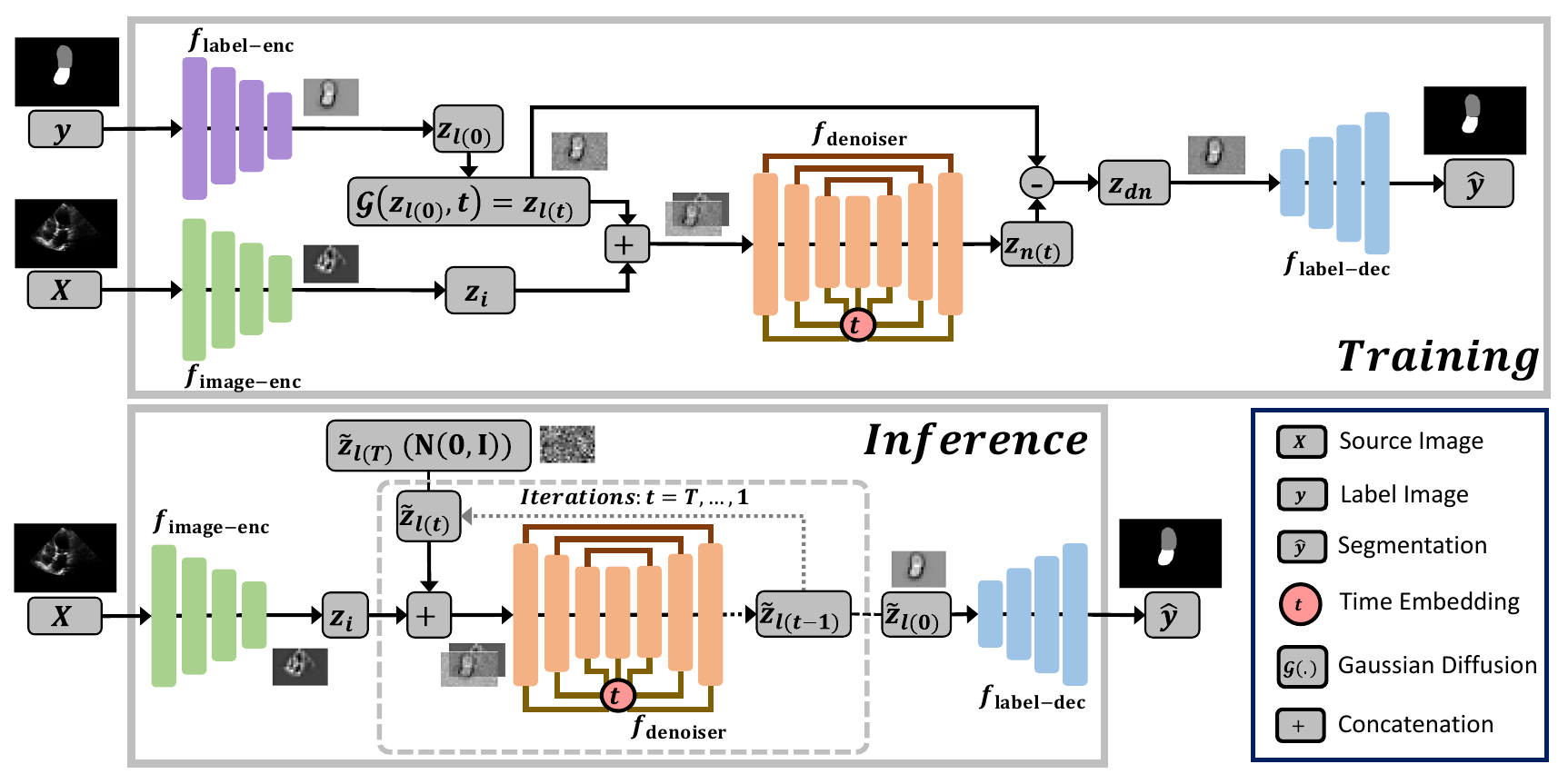}
    \caption{The proposed LDSeg model. The label encoder $f_\text{label-enc}$ and image encoder $f_\text{image-enc}$ are used to obtain corresponding low dimensional latent representations $z_{l(0)}$ and $z_i$ for a given ground truth label/mask image $y$ and source image $X$, respectively. A denoiser $f_\text{denoiser}$, conditioned on the source image embedding $z_i$, is used to learn the noise distributions of $z_{l(t)}$ for timesteps $t=1, \dotsc ,T$, where $T$ is the total number of diffusion steps. $z_{l(t)}$ is obtained by perturbing $z_{l(0)}$ with a Gaussian block $\mathcal G(\cdot)$ for a given noise variance scheduler $\alpha$ and $\beta$. The cleaned latent space $z_{dn}$ is obtained by subtracting the predicted noise $z_{n(t)}$ from the perturbed one $z_{l(t)}$. Finally, a label decoder $f_{\text{label-dec}}$ is used to obtain the segmentation $\hat{y}$ of the semantic labels in the original image from $z_{dn}$. The model is trained in an end-to-end fashion, where our objective is to learn $q(\hat{y}|X)=\mathbb{E}_{q_{i}(z_{i}|X)}\left[q_{s}(\hat{y}|z)\right]$, where $q_{l}(z \mid y, X) \sim \mathcal{N}(z_{dn}, \sigma^2 \mathrm{I})$. In the inference phase, starting with a random Gaussian $\tilde{z}_{l(T)} \sim \mathcal{N}(\mathrm{0,I})$, the denoiser is iterated for timestep $t=T, \dotsc ,1$ to obtain $\tilde{z}_{l(0)}$ with $z_i$ as the condition. Final segmentation $\hat{y}=f_{\text{label-dec}}(\tilde{z}_{l(0)})$ is obtained using the trained label decoder.}
    \label{Figure1}
\end{figure*}

The choice of Gaussian provides a close-form solution to generate a transitional state $x_t$ using, 
\begin{equation}
    x_t = \sqrt{\bar{\alpha}}x_0 + \sqrt{1-\bar{\alpha}}\epsilon
\end{equation}
where $\alpha_t=1-\beta_t$, $\bar{\alpha_t}=\prod_{i=1}^{t}\alpha_i$ and $\epsilon \sim \mathcal{N}(\mathrm{0,I})$. Training is usually performed by optimizing the variational bound on the negative log likelihood of $p_{\theta}(x_0)$:

\begin{equation}
    \mathcal{L} \coloneqq \mathbb{E}_{q} \left[ -\log \frac{p_{\theta}(x_{0:T})}{q(x_{1:T} \mid x_0)} \right] \geq \mathbb{E} \left[-\log p_{\theta}(x_0) \right]
\end{equation}
However, with re-parameterization, Ho {\em et al.} \cite{ho2020denoising} simplified the training objective by proposing a variant of the variational bound that improves the quality of generated samples while being easier to implement,   
\begin{equation}
    \mathcal{L}_{DDPM} \coloneqq \mathbb{E}_{t,x_0,\epsilon}[\Vert \epsilon - \epsilon_{\theta}(x_t,t) \Vert^2]
\end{equation}
where $\epsilon_{\theta}$ is a function approximator intended to predict $\epsilon$ from $x_t$ by a trained denoiser. With a trained denoiser, the data can be generated with the reverse process by iterating through $t=T, \dotsc ,1$. Starting from $x_T \sim \mathcal{N}(\mathrm{0,I})$, the transitional states can be obtained by,
\begin{equation}
    x_{t-1} = \frac{1}{\sqrt{\alpha_{t}}}\left(x_t - \frac{\beta_t}{\sqrt{1-\bar{\alpha_t}}} \epsilon_{\theta}(x_t,t)\right) + \sigma_{t}z
\end{equation}
where $\sigma_t$ is the noise variance of timestep $t$ and $z \sim \mathcal{N}(\mathrm{0,I})$.

\section{Method}

The proposed LDSeg framework consists of four major components: 
\begin{enumerate}
\item {\em Label encoder:} The label encoder, denoted by $f_{\rm label-enc}$, is used to learn the low-dimensional latent representation with a standardized distribution of the shape manifolds of the target object. 
\item {\em Image encoder:} The image encoder, denoted by $f_{\rm image-enc}$, learns the low-dimensional image embedding $z_i$ from the source image $X$. 
\item {\em Conditional label denoiser:} The denoiser $f_{\rm denoiser}$ learns the added noise of the perturbed label embedding for each time step, conditioned on the embedding of the source image and the time step $t$. 
\item {\em Label decoder:} The label decoder $f_{\rm label-dec}$ is used to produce segmentation by mapping the denoised latent space to its corresponding semantic label image in the original image domain. The model training and inference workflows are shown in Figure \ref{Figure1}.
\end{enumerate}

\begin{figure*}[ht]
    \centering
    \includegraphics[width=7in]{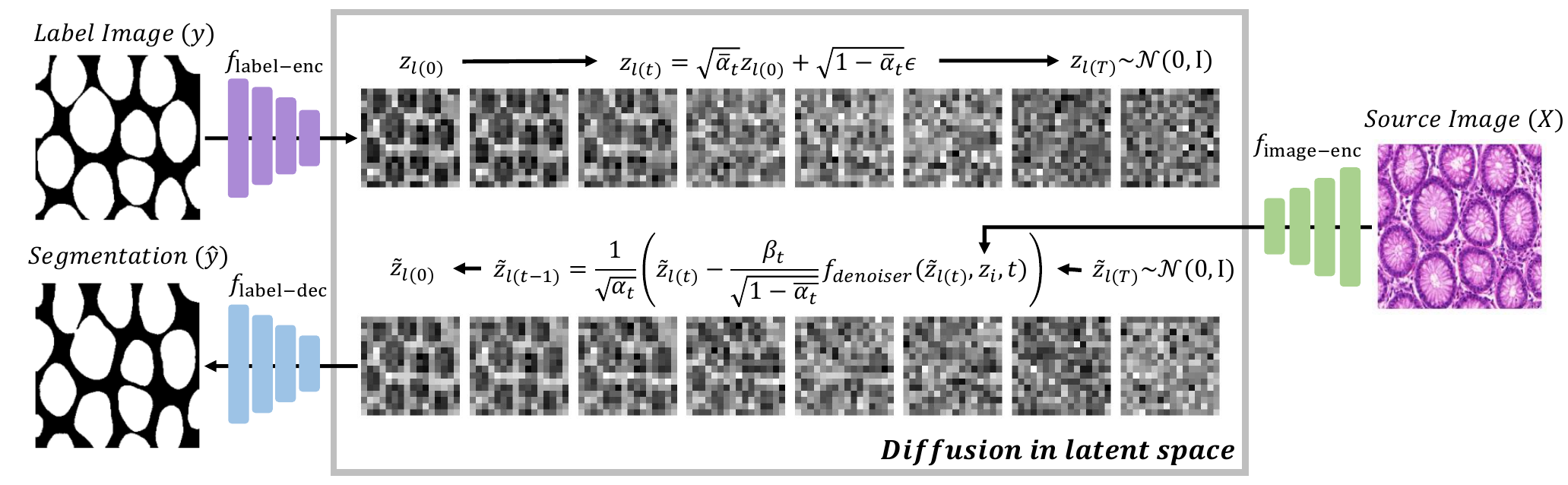}
    \caption{A sample GlaS data \cite{sirinukunwattana2016gland} was used to demonstrate the forward and the reverse diffusion processes. In the forward process (top row), the low-dimensional latent representation $z_{l(0)}$ is first obtained from the label image. Then, Gaussian noise is gradually injected for timestep $t=1, \dotsc ,T$, given the noise variance schedules of $\beta$, where $\epsilon \sim \mathcal{N}(\mathrm{0,I})$. At timestep $T$, $z_{l(T)}$ is subject to $\mathcal{N}(\mathrm{0,I})$. To start the reverse process (bottom row), $\tilde{z}_{l(T)}$ is sampled from $\mathcal{N}(\mathrm{0,I})$. Then the denoiser is used iteratively for timesteps $t=T, \dotsc, 1$ with the source image embedding $z_i$ as the condition. At the end of the reverse process, the segmentation mask is obtained from $\tilde{z}_{l(0)}$ using the trained label decoder.}
    \label{Figure2}
\end{figure*}

We note that the segmentation labels are discrete, and hence corrupting them by Gaussian noise is unnatural, as the label/mask image has only a few modes (i.e., the number of object classes). We propose to mitigate this inherent problem by learning a low-dimensional standardized representation of the label images. In other words, we want to learn a label encoder $f_\text{label-enc}(\cdot)$ that projects the input labels into a latent space with standardized distribution. Essentially, the label encoder learns to produce low-dimensional latent representation of the object shape manifolds for the label images. This low-dimensional standardized representation (label embedding) has two major advantages over the original label image, (1) it is continuous, thus ensuring smooth transition among different object classes, and (2) it is computationally more efficient to train a conditional denoiser for a low-dimensional standardized latent space, thus making the algorithm significantly faster in the inference phase.

A standard DPM denoiser has two inputs, a noisy version of the input image and its corresponding timestep. For segmentation, the denoiser needs additional conditioning. The condition can be the source image \cite{bogensperger2023scorebased,wu2022medsegdiff}, or a text indicating the target object \cite{ldznet}. As our objective is image semantic segmentation, we propose to use image embedding as a condition for the denoiser. The image embedding is a low-dimensional latent representation of the source image having the same size as the label embedding, which is learned using an image encoder $f_\text{image-enc}(\cdot)$. The image embedding is concatenated with the noisy representation of the label embedding and used as a two channel input to the denoiser, along with its corresponding timestep as a separate input. The denoiser $f_\text{denoiser}(\cdot)$, learns the transitional noisy distributions of the label embedding, conditioned on the image embedding, and predicts noise for a given timestep. Finally, to map the denoised latent space to the semantic segmentation in the original image domain, we learn a label decoder $f_\text{label-dec}$.

\subsection{Loss function}
Let, $X$, $y$, and $\hat{y}$ be the source image, its corresponding label image, and the predicted segmentation, respectively, sampled from the dataset. $z_i$, $z_{l(0)}$ are their corresponding image embedding and label embedding,
\begin{align}
    z_i &= f_{\text{image-enc}}(X) \\[10pt]
    z_{l(0)} &= f_{\text{label-enc}}(y)
\end{align}

Given noise variance schedule parameters $\alpha$ and $\beta$, a Gaussian block $\mathcal G(\cdot)$ is used to produce the noisy $z_{l(t)}$ for timestep $t \in (1, T)$ \cite{ho2020denoising, nichol2021improved},
\begin{equation}
    z_{l(t)} = \mathcal G(z_{l(0)}, t) = \sqrt{\bar{\alpha_{t}}}z_{l(0)}+\sqrt{1-\bar{\alpha_{t}}}\epsilon
\end{equation}
where $\alpha_t=1-\beta_t$, $\bar{\alpha_t}=\prod_{i=1}^{t}\alpha_i$ and $\epsilon \sim \mathcal{N}(\mathrm{0,I})$. The denoiser predicts the noise of the timestep $t$ conditioned on $z_i$. Let $z_{dn}$ be the denoised latent space,
\begin{align}
    z_{dn} &= z_{l(t)} - f_{\text{denoiser}}(z_{l(t)}, z_i, t) \\[10pt]
    \hat{y} &= f_{\text{label-dec}}(z_{dn})
\end{align}

Our objective is to learn $q(\hat{y}|X)=\mathbb{E}_{q_{i}(z_{i}|X)}\left[q_{s}(\hat{y}|z)\right]$, where $q_{l}(z \mid y, X) \sim \mathcal{N}(z_{dn}, \sigma^2 \mathrm{I})$. The loss function $\mathcal{L}$ consists of two terms, the segmentation loss $\mathcal{L}_{1}$ and the denoiser loss $\mathcal{L}_{2}$, where the segmentation loss is a combination of cross-entropy loss $\mathcal{L}_{CE}$ and dice similarity coefficient (DSC) loss $\mathcal{L}_{DSC}$.
\begin{align}
    \mathcal{L}_{\text{CE}}(\hat{y}, y) &= -\sum_{c \in C} y_c \log(\hat{y}_c) \\[10pt]
    \mathcal{L}_{\text{DSC}}(\hat{y}, y) &= 1 - \frac{2 \sum_i \hat{y}_i y_i}{\sum_i \hat{y}_i + \sum_i y_i} \\[10pt]
    \mathcal{L}_{1} &= \mathbb{E}_{X, y} \left[ \mathcal{L}_{\text{CE}}(\hat{y}, y) + \gamma \mathcal{L}_{\text{DSC}}(\hat{y}, y) \right] \\[10pt]
    \mathcal{L}_{2} &= \mathbb{E}_{\epsilon \sim \mathcal{N}\mathrm{(0, I)}} \left[ \| f_{\text{denoiser}}(z_{l(t)}, z_i, t) - \epsilon \|^2 \right] \\[10pt]
    \mathcal{L} &= \mathcal{L}_{1} + \lambda \mathcal{L}_{2}
\end{align}
where, $c \in C$ is an object class of a set of object classes $C$, $i$ is the corresponding pixel, $\gamma$ and $\lambda$ are scaler co-efficients. This end-to-end training strategy is functionally analogous to a VAE model. Although it does not explicitly parameterize a distribution like VAEs, the denoiser's role in handling noise in latent space creates an analogous structure. Unlike VAE, the proposed model does not aim to explicitly match a latent distribution to a priori. Instead, the denoiser's regularization ensures that the latent space remains structured, noise-resilient and better representative of the segmentation related features.

An example of a conditional denoising forward process for segmentation generation is shown in Figure \ref{Figure2} (top row). The algorithm for end-to-end training is shown in Algorithm \ref{alg:training}.

\begin{figure}[t]
\begin{algorithm}[H]
  \caption{Training} \label{alg:training}
  \begin{algorithmic}[1]
    \Repeat
      \State $X,y \sim q_{data}(X,y)$
      \State $z_i = f_{\text{image-enc}}(X)$
      \State $z_{l(0)} = f_{\text{label-enc}}(y)$
      \State $t \sim \mathrm{Uniform}(\{1, \dotsc, T\})$
      \State $\epsilon \sim \mathcal{N}(\mathrm{0,I})$
      \State $z_{l(t)} = \sqrt{\bar{\alpha_{t}}}z_{l(0)}+\sqrt{1-\bar{\alpha_{t}}}\epsilon$
      \State $z_{dn} = z_{l(t)} - f_{\text{denoiser}}(z_{l(t)}, z_i, t)$
      \State $\hat{y} = f_{\text{label-dec}}(z_{dn})$
      \State Take gradient descent step on
      \vspace{5pt}
      \Statex \footnotesize $\; \quad \nabla \left[ \left[ \mathcal{L}_{\text{CE}}(\hat{y}, y) + \gamma \mathcal{L}_{\text{DSC}}(\hat{y}, y) \right] + \lambda \left\| \epsilon - f_{\text{denoiser}}(z_{l(t)}, z_i, t) \right\|^2 \right]$ \normalsize
      \vspace{5pt}
    \Until{converged}
  \end{algorithmic}
\end{algorithm}
\end{figure}

\subsection{Reverse Process for Segmentation}
As the image encoder is independent of the denoiser, we only need to obtain the image embedding $z_i$ at the start of the reverse process. In the reverse process, the main objective is to generate latent representation $z_{l(0)}$, conditioned on $z_i$. Like the other image generation tasks of DPMs, a Gaussian $\mathcal{N}(\mathrm{0,I})$ is used as the noisy latent mask representation $\tilde{z}_{l(T)}$ at timestep $T$. Then the denoiser is iterated for $t = T, \dotsc ,1$. At the end of the iteration, we obtain $\tilde{z}_{l(0)}$, which is used as an input to the trained label decoder to get the final segmentation $\hat{y} = f_{\text{label-dec}}(\tilde{z}_{l(0)})$. An example of reverse process is shown in Figure \ref{Figure2} (bottom row). The sampling algorithm for segmentation is shown in Algorithm \ref{alg:inference}.

\begin{figure}[ht]
\begin{algorithm}[H]
  \caption{Inference} \label{alg:inference}
  \begin{algorithmic}[1]
    \State $X \sim q_{data}(X)$, $\tilde{z}_{l(t)} \sim \mathcal{N}(\mathrm{0,I})$
    \State $z_i = f_{\text{image-enc}}(X)$
    \For{$t=T, \dotsc, 1$}
      \State $n \sim \mathcal{N}(\mathrm{0,I})$ if $t > 1$, else $n = 0$
      \vspace{5pt}
      \State \footnotesize $\tilde{z}_{l(t-1)} = \frac{1}{\sqrt{\alpha_t}}\left( \tilde{z}_{l(t)} - \frac{\beta_t}{\sqrt{1-\bar\alpha_t}} f_{\text{denoiser}}(\tilde{z}_{l(t)}, z_i, t) \right) + \sigma_t n$ \normalsize
      \vspace{5pt} 
    \EndFor
    \State $\hat{y} = f_{\text{label-dec}}(\tilde{z}_{l(0)})$
    \State \textbf{return} $\hat{y}$
  \end{algorithmic}
\end{algorithm}
\end{figure}

\section{Experiments}

\subsection{Datasets}
We have used 3 datasets to demonstrate the effectiveness of the proposed LDSeg:
\begin{enumerate}
    \item \textbf{Echo} is a 2D+time echocardiogram (echo) video dataset from University of Iowa Hospitals \& Clinics. All the videos are standard apical 4-chamber scans with a left-ventricular focused view. Echos are acquired by transthoracic echocardiography (TTE) using standard 2D echocardiography techniques following the guidelines of the American Society of Echocardiography. In total, the dataset contains 65 echos (2230 still frames). The left ventricles (LV) and the left atria (LA) were fully traced by an expert manually using ITK-snap.
    \item \textbf{GlaS} \cite{sirinukunwattana2016gland} is a publicly available 2D histopathology dataset of Hematoxylin and Eosin (H\&E) stained slides, acquired by a team of pathologists at the University Hospitals Coventry and Warwickshire, UK. The training set contains 37 benign and 48 malignant cases, whereas the test set contains additional 37 benign and 43 malignant cases.
    \begin{figure}[ht!]
        \centering
        \includegraphics[width=0.32\textwidth]{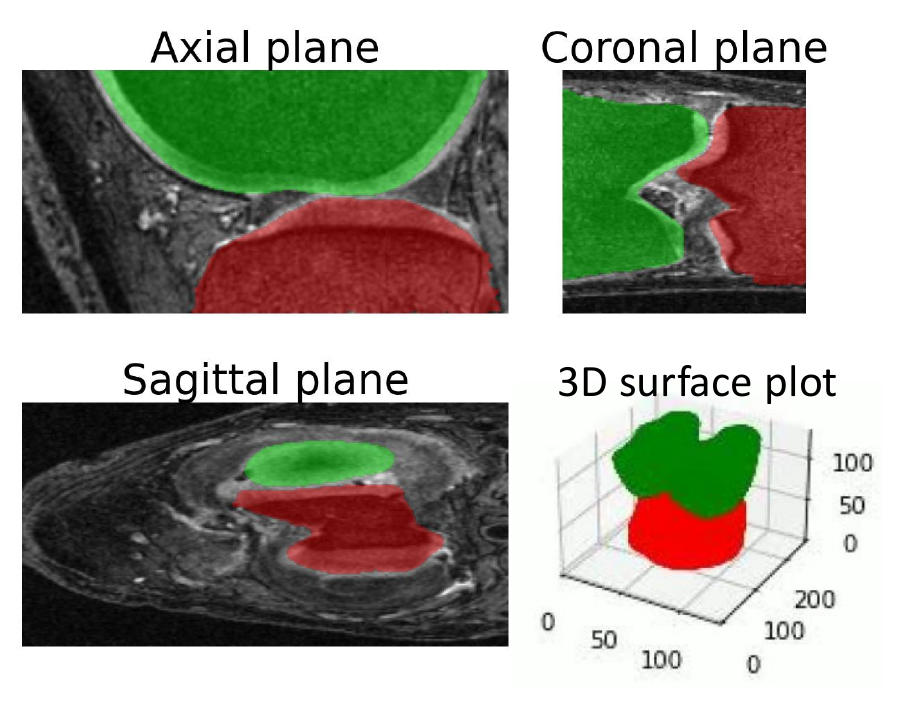}
        \caption{A sample of knee data. FC and TC are marked with green and red color. Three slices from axial, coronial and sagittal plane is shown along with the 3D surface plot for FC and TC.}
        \label{Figure3}
    \end{figure}
    \item \textbf{Knee} (\href{https://data-archive.nimh.nih.gov/oai/}{https://data-archive.nimh.nih.gov/oai/}) is a publicly available 3D MRI dataset. The dataset contains randomly selected 987 3D MRI scans from 244 patients on different time points. Focused volumetric regions with an image size of $160 \times 104 \times 256$ around the femur cartilage with bone (FC) and tibia cartilage with bone (TC) are used as the region of interest (ROI). The FC and TC are segmented by an automatic segmentation algorithm and validated/edited by an expert. Figure \ref{Figure3} shows a sample of the Knee dataset.
\end{enumerate}

\begin{figure*}[ht!]
    \centering
    \includegraphics[width=7in]{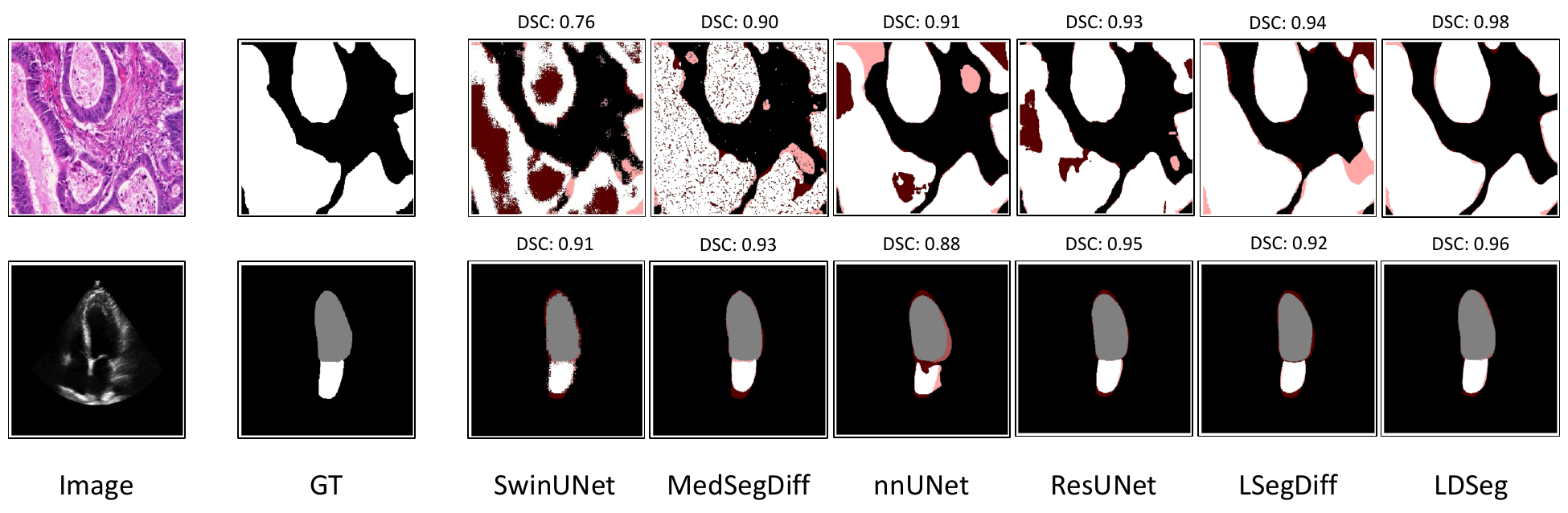}
    \caption{Qualitative segmentation results of different methods for GlaS and Echo dataset, shown in top and bottom rows, respectively. Dark red marks the false negative and the light red marks the false positive error on the segmentation result. GT indicates the ground-truth/label-image.}
    \label{Figure4}
\end{figure*}

\subsection{Model Architecture}
The label and image encoder both have architectures similar to standard ResUnet encoder \cite{zhang2018road}, without any skip connections. Each have several convolution and down-sampling layers that determine the size of the latent space for the low-dimensional projection of the label and source input images. We experimented with different down-sampling scales and chose $4$ down-sampling layers, which produced the best results for all three datasets. The image size for Echo, GlaS and Knee data were resized to $512 \times 768$, $512 \times 512$ and $128 \times 128 \times 256$, respectively. Hence, the sizes of the low-dimensional $z_{l(0)}$ and $z_i$ for Echo, GlaS, and Knee data are $32 \times 48$, $32 \times 32$ and $8 \times 8 \times 16$, respectively. We observed that these were the optimal latent sizes as further down-sampling reduced the model accuracy, while less down-sampling reduced denoiser accuracy as the search space got enlarged and learning noise distributions became challenging. A normalization layer is added as the final layer to the label encoder to ensure standardized distribution ($\mu=0, \sigma=1$) for label embedding. On the other hand, final two down-sampling layers of the image encoder are equipped with multi-head attention layers \cite{cordonnier2020multi} to capture robust imaging features. The denoiser has a standard ResUnet shape with time-embedding blocks and self-attention layers. Specifically, we have adapted the denoiser architecture from \cite {ho2020denoising}. The image embedding is concatenated with the noisy representation of the label embedding and used as a two channel input to the denoiser, along with it's corresponding timestep as a separate input. The decoder have similar architecture to the decoder of ResUnet, without the skip connections. A softmax activation layer is used as the final layer of the decoder to obtain the probabilistic distribution of different object classes. 

\subsection{Experimental Setup}
Exponentially decayed learning rates were used to train the models with $1\times10^{-2}$ and $1\times10^{-3}$ as the initial learning rates for all the model components. For the Echo and Knee datasets, we used a $80\%:20\%$ split for training and testing, and among each training dataset, $10\%$ were used for validation during model development. The well-separated training and test sets were used for the GlaS dataset~\cite{sirinukunwattana2016gland}. The noise step $t$ is an integer randomly sampled from $1$ to $1000$ for each batch. For the loss function, choices of $\gamma=2$ and $\lambda=1$ gave the best model accuracy. NVIDIA A100-SXM4 (80GB) GPU was used for model training and inference.

\section{Results}

\subsection{Segmentation Accuracy}
\begin{table}[ht]
    \centering
    \caption{Quantitative results for Echo data segmentation. }
    \label{table-1}
    \small
    \setlength{\tabcolsep}{0.5em}
    \renewcommand{\arraystretch}{1.2}
    \resizebox{0.45\textwidth}{!}{
    \begin{tabular}{c|ccc|ccc}
        \hline
		\multirow{2}{*}{\textbf{Method}} & \multicolumn{3}{c|}{\textbf{DSC \textsuperscript{$\uparrow$}}} & \multicolumn{3}{c}{\textbf{IoU \textsuperscript{$\uparrow$}}} \\
        \cline{2-7}
        {} & LV & LA & LV+LA & LV & LA & LV+LA \\
  		\hline
        SwinUNet \cite{cao2022swin} & $0.85$ & $0.70$ & $0.81$ & $0.75$ & $0.56$ & $0.69$ \\
		U-net \cite{ronneberger2015u} & $0.86$ & $0.75$ & $0.83$ & $0.77$ & $0.62$ & $0.72$ \\
        nnUNet \cite{isensee2021nnu} & $0.89$ & $0.76$ & $0.86$ & $0.81$ & $0.64$ & $0.76$ \\
        MedSegDiff\textsuperscript{ *} \cite{wu2022medsegdiff} & $0.89$ & $0.81$ & $0.87$ & $0.82$ & $0.70$ & $0.78$ \\
        V-net \cite{milletari2016v} & $\mathbf{0.93}$ & $0.81$ & $0.90$ & $\mathbf{0.87}$ & $0.71$ & $0.83$ \\
        Res-Unet \cite{zhang2018road} & $\mathbf{0.93}$ & $0.83$ & $0.91$ & $\mathbf{0.87}$ & $0.74$ & $\mathbf{0.84}$ \\
        LSegDiff\textsuperscript{*} \cite{vu2023lsegdiff} & $\mathbf{0.93}$ & $0.85$ & $0.91$ & $\mathbf{0.87}$ & $0.75$ & $\mathbf{0.84}$ \\
        LDSeg\textsuperscript{*} (Ours) & $\mathbf{0.93}$ & $\mathbf{0.87}$ & $\mathbf{0.92}$ & $\mathbf{0.87}$ & $\mathbf{0.77}$ & $\mathbf{0.84}$ \\
        \hline
        \multicolumn{7}{l}{\footnotesize * denotes the DPMs.}
    \end{tabular}}
\end{table}

\begin{table}[ht]
    \centering
    \caption{Quantitative results for GlaS data segmentation. }
    \label{table-2}
    \small
    \setlength{\tabcolsep}{0.5em}
    \renewcommand{\arraystretch}{1.2}
    \resizebox{0.3\textwidth}{!}{
    \begin{tabular}{c|c|c}
        \hline
		\textbf{Method} & \textbf{DSC \textsuperscript{$\uparrow$}} & \textbf{IoU \textsuperscript{$\uparrow$}} \\
        \hline
        SwinUNet \cite{cao2022swin} & $0.76$ & $0.62$ \\
		U-net \cite{ronneberger2015u} & $0.78$ & $0.65$ \\
        U-net++ \cite{zhou2018unet++} & $0.78$ & $0.66$ \\
        MedT \cite{valanarasu2021medical} & 0.81 & 0.70 \\
        SDF-DDPM\textsuperscript{ *} \cite{bogensperger2023scorebased} & $0.83$ & $0.72$ \\
        nnUNet \cite{isensee2021nnu} & $0.84$ & $0.73$ \\
        LSegDiff\textsuperscript{ *} \cite{vu2023lsegdiff} & $0.84$ & $0.74$ \\
        MedSegDiff\textsuperscript{ *} \cite{wu2022medsegdiff} & $0.84$ & $0.74$ \\
        Res-Unet \cite{zhang2018road} & $0.86$ & $0.76$ \\
        LDSeg\textsuperscript{*} (Ours) & $\mathbf{0.91}$ & $\mathbf{0.84}$ \\
        \hline
        \multicolumn{3}{l}{\footnotesize * denotes the DPMs.}
    \end{tabular}}
\end{table}

\begin{table}[ht]
    \centering
    \caption{Quantitative results for the Knee data segmentation. }
    \label{table-3}
    \small
    \setlength{\tabcolsep}{0.5em}
    \renewcommand{\arraystretch}{1.2}
    \resizebox{0.45\textwidth}{!}{
    \begin{tabular}{c|ccc|ccc}
        \hline
		\multirow{2}{*}{\textbf{Method}} & \multicolumn{3}{c|}{\textbf{DSC \textsuperscript{$\uparrow$}}} & \multicolumn{3}{c}{\textbf{IoU \textsuperscript{$\uparrow$}}} \\
        \cline{2-7}
        {} & FC & TC & FC+TC & FC & TC & FC+TC \\
  		\hline
        SwinUNet \cite{cao2022swin} & $0.85$ & $0.70$ & $0.81$ & $0.75$ & $0.56$ & $0.69$ \\
        LSegDiff\textsuperscript{*} \cite{vu2023lsegdiff} & $0.96$ & $\mathbf{0.96}$ & $\mathbf{0.96}$ & $\mathbf{0.93}$ & $0.92$ & $\mathbf{0.93}$ \\
        Res-Unet \cite{zhang2018road} & $\mathbf{0.97}$ & $\mathbf{0.96}$ & $\mathbf{0.96}$ & $\mathbf{0.93}$ & $0.93$ & $\mathbf{0.93}$ \\
        nnUNet \cite{isensee2021nnu} & $\mathbf{0.97}$ & $\mathbf{0.96}$ & $\mathbf{0.96}$ & $\mathbf{0.93}$ & $\mathbf{0.94}$ & $\mathbf{0.93}$ \\
        LDSeg\textsuperscript{*} (Ours) & $0.96$ & $\mathbf{0.96}$ & $\mathbf{0.96}$ & $\mathbf{0.93}$ & $0.92$ & $\mathbf{0.93}$ \\
        \hline
        \multicolumn{7}{l}{\textsuperscript{$\star$} denotes the DPMs.}
    \end{tabular}}
\end{table}

We evaluated the performance of our proposed method LDSeg using two standard metrics: (1) Dice Similarity Co-efficient (DSC) and (2) Intersection over Union (IoU). Tables \ref{table-1},\ref{table-2} and \ref{table-3}, show the quantitative results for different methods for Echo, GlaS, and Knee datasets, respectively. Figure \ref{Figure4} shows qualitative segmentation results by different methods of GlaS and Echo dataset. LDSeg achieves the best DSC and IoU scores for all the datasets. SDF-DDPM method uses signed distance functions to represent mask images, which is ambiguous for data with multi-labels. Hence, it is only reported for the GlaS dataset. For the 3D Knee dataset, it was impossible to implement the full architecture of MedSegDiff due to the large GPU memory consumption. This indicates that diffusions in the latent space is of tremendous help for segmenting 3D medical images with a large image size when the GPU/CPU memory is constrained.

\subsection{Computational Efficiency}
\begin{figure}[ht!]
    \centering
    \includegraphics[width=0.4\textwidth]{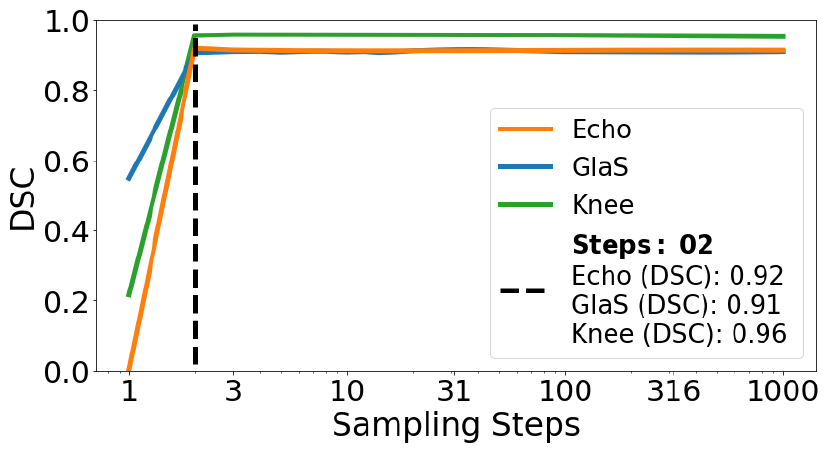}
    \caption{The number of evenly spaced sampling steps vs DSC for different datasets. The DDIM algorithm with only $2$ evenly spaced sampling steps between $1$ and $T=1000$ (inclusive) produced maximum segmentation accuracy for all the datasets. The number of steps are plotted in the logarithmic scale for convenience.}
    \label{Figure5}
\end{figure}
\begin{table}[ht]
    \centering
    \caption{The LDSeg algorithm run times for segmenting a single image from each dataset are shown for sampling steps $1000$ (using all the sampling steps) and sampling steps $2$ (the minimum number of sampling steps to achieve the same accuracy as using all sampling steps).}
    \label{table-4}
    \small
    \setlength{\tabcolsep}{0.5em}
    \renewcommand{\arraystretch}{1.5}
    \resizebox{0.4\textwidth}{!}{
    \begin{tabular}{c|c|c}
        \hline
		\multirow{2}{*}{\textbf{Dataset}} & \multicolumn{2}{c}{\textbf{Execution time (seconds)}} \\
        \cline{2-3}
        {} & {Sampling steps=$1000$} & {Sampling steps=$2$} \\
        \hline
		Echo & $91.23$ & $0.34$ \\
        GlaS & $80.37$ & $0.30$ \\
        Knee & $132.36$ & $0.49$ \\
        \hline
    \end{tabular}}
\end{table}
\begin{figure}[ht]
    \centering
    \includegraphics[width=0.37\textwidth]{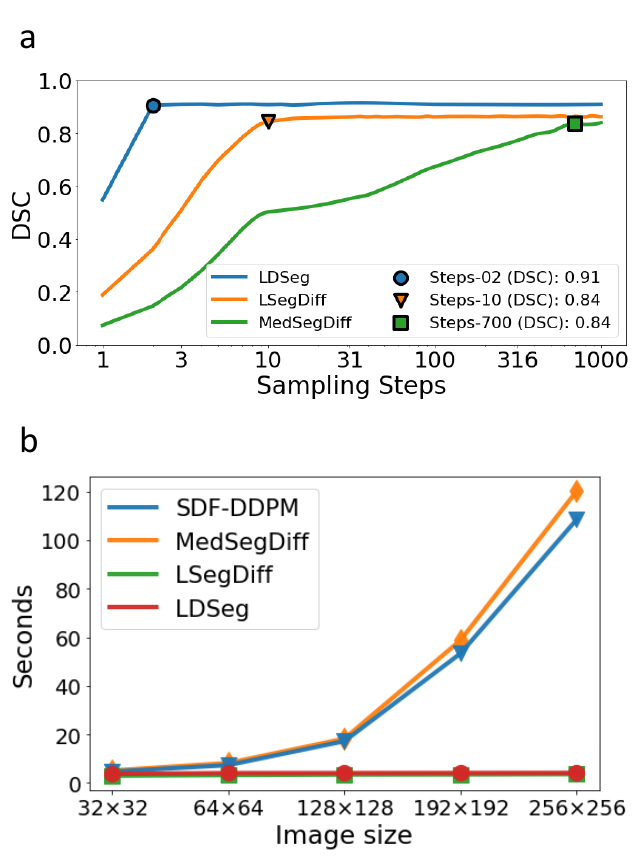}
    \caption{\textbf{(a)} The number of sampling steps vs DSCs using LDSeg, LSegDiff and MedSegDiff models for the GlaS dataset. LDSeg was able to achieve the maximum DSC with only $2$ sampling steps, outperforming both LSegDiff ($10$) and MedSegDiff ($700$). \textbf{(b)} Image sizes vs execution times for segmenting a single image with different DPM. The execution times of LDSeg and LSegDiff (both latent diffusion models) remained close to constant due to the use of constrained low-dimensional latent space, while for SDF-DDPM and MedSegDiff, the execution times increased exponentially with increased image sizes.}
    \label{Figure6}
\end{figure}
\begin{figure*}[ht]
    \centering
    \includegraphics[width=5in]{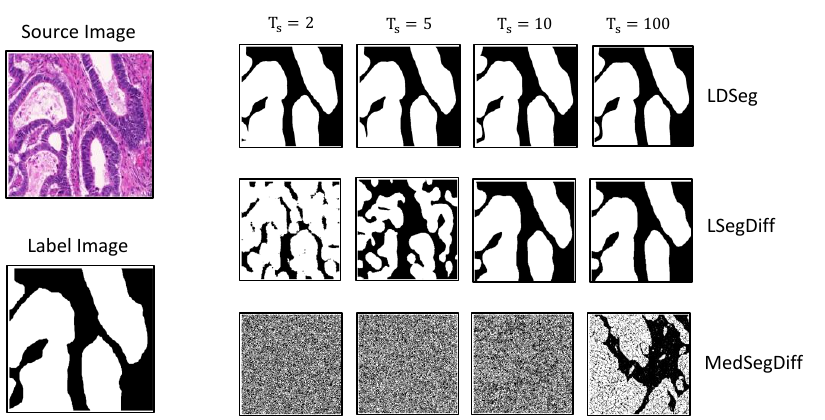}
    \caption{Qualitative comparison of the segmentations results for various sampling steps ($T_s$) of an image from GlaS dataset for different methods. LDSeg produced qualitatively good segmentation even with $2$ sampling steps, whereas LSegDiff needed atleast $10$ (both are latent diffusion models). MedSegDiff, which performs diffusion in the original image domain, could not produce reasonably good segmentation even with $100$ sampling steps.}
    \label{Figure7}
\end{figure*}
\begin{figure*}[ht]
    \centering
    \includegraphics[width=7in]{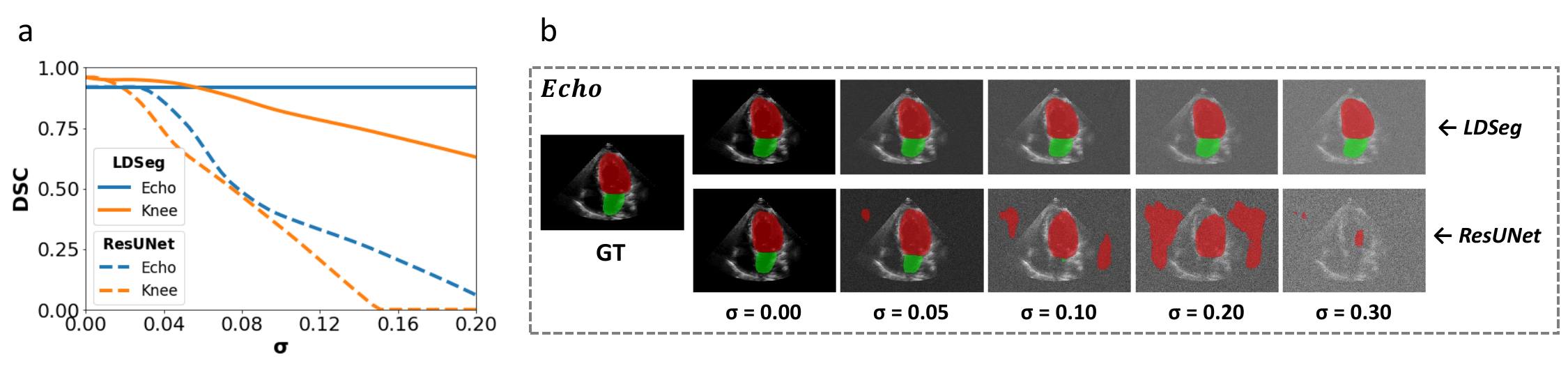}
    \caption{\textbf{(a)} Added noise variance $\sigma$ versus DSC scores for ResUnet and LDSeg on the Echo and Knee dataset. LDSeg (solid lines) significantly outperformed ResUNet model (dotted lines), which is a deterministic model, in terms of noise resilience on the source image. \textbf{(b)} The top and bottom rows show some sample segmentation results for an Echo data for noisy images with different noise variances, using LDSeg and ResUNet, respectively. GT indicates the ground truth of the image.}
    \label{Figure8}
\end{figure*}

The major difference of LDSeg compared to other traditional diffusion-based segmentation methods is that the diffusion operations are performed in the latent low-dimensional space. We expect the memory demand and total sampling time of LDSeg for a sampling sequence to be less than the other diffusion-based methods. We further experimented on the sampling sequence for the reverse process. Nichol {\em et al.} \cite{nichol2021improved} observed that the model trained with the $``cosine"$ noise scheduler, and sampled with DDIM algorithm \cite{song2020denoising} performed remarkably well in image generation. DDIM method proposed by Song {\em et al.} deterministically maps noises to images without using stochasticity in the transition states. Nichol {\em et al.} used fewer sampling steps $(<50)$, when $T>=1000$ and achieved close to the optimal Fréchet inception distance (FID) score for image generation. They used $K$ evenly spaced real numbers between $1$ and $T$ (inclusive) as sampling steps, and then rounded each resulting number to the nearest integer value. We adapted the same sampling strategy and observed that with remarkably less sampling steps (sampling steps $=2$), LDSeg is able to achieve the same segmentation accuracy as using all the sampling steps, for all the test datasets. Figure \ref{Figure5} shows the number of sampling steps versus the DSC scores for all the datasets with the DDIM sampling algorithm. Table \ref{table-4} shows the execution times (in seconds) needed to segment a single image using all the sampling steps for $T=1000$, and the minimum number of sampling steps needed to achieve the same segmentation accuracy. With the optimal number of sampling steps, LDSeg achieved a significant increase in the efficiency of the sampling time ($\sim 268$ times reduction in the execution time).

We further investigated the execution times for segmenting a single image with different image sizes for different DPMs. The minimum number of sampling steps to achieve maximum segmentation accuracy (i.e., achieved using all sampling steps) can be different (Figure \ref{Figure6}(a)) for different DPM due to different objectives to learn target noise distributions. For a fair comparison, we fixed the total sampling steps to $50$ and performed the experiments on the GlaS dataset by down-sampling the images into different sizes. Figure \ref{Figure6}(b) shows that with increased image sizes, the execution times for SDF-DDPM and MedSegDiff increased exponentially, mainly due to denoising in the higher-dimensional image domain. For LDSeg, the execution times were close to constant with the increased image sizes, as the denoising in the low-dimensional latent space is efficient and less time/memory consuming. The improvement of LDSeg over LSegdiff, both in terms of computation time and performance can be attributed to the end-to-end training strategy. In particular, the use of the MSE loss in the latent domain does not necessarily capture the segmentation errors, compromising the performance. Figure \ref{Figure7} shows the qualitative comparison of the segmentation results for LDSeg, LSegDiff and MedSegDiff methods for an image of GlaS dataset, using different sampling steps. With only $2$ sampling steps, LDSeg is comparable to deterministic models like ResUnet for image segmentation in terms of computation time.

\subsection{Robustness to Noise}
\begin{table*}[ht]
    \centering
    \caption{GlaS Ablation study. }
    \label{table-5}
    \small
    \setlength{\tabcolsep}{0.5em}
    \renewcommand{\arraystretch}{1.2}
    \resizebox{0.6\textwidth}{!}{
    \begin{tabular}{c|cccc|cc}
        \hline
		\multirow{2}{*}{\textbf{Method}} & \textbf{Label} & \textbf{Image} & \textbf{Label} & \textbf{Image} & \multirow{2}{*}{\textbf{DSC}} & \multirow{2}{*}{\textbf{IoU}} \\
        {} & \textbf{Encoder} & \textbf{Encoder} & \textbf{Down-sample} & \textbf{Down-sample} & {} & {} \\
  		\hline
		LDSeg\textsubscript{(ld,id)} & \ding{56} & \ding{56} & \ding{52} & \ding{52} & $0.47$ & $0.33$ \\
        LDSeg\textsubscript{(id)} & \ding{52} & \ding{56} & \ding{56} & \ding{52} & $0.61$ & $0.47$ \\
        LDSeg\textsubscript{(ld)} & \ding{56} & \ding{52} & \ding{52} & \ding{56} & $0.71$ & $0.56$ \\
        LDSeg & \ding{52} & \ding{52} & \ding{56} & \ding{56} & $\mathbf{0.91}$ & $\mathbf{0.84}$ \\
        \hline
        \multicolumn{7}{c}{\normalsize ld $\rightarrow$ Label Down-sampled, id $\rightarrow$ Image Down-sampled.} \\
    \end{tabular}}
\end{table*}
One of the key challenges for medical image segmentation is to produce accurate segmentation from noisy image acquisition. Often times, deterministic segmentation models fail in the presence of noise in the test dataset. As the denoiser in LDSeg is conditioned on the source image embedding, which is a low-dimensional representation of the source image, intuitively it should be more robust to high-frequency noise. Moreover, the learned shape manifolds of the target object act as priori to the denoiser for iterative denoising even with the noisy or slightly inaccurate image embedding, and helps produce clean latent representation $\tilde{z}_{l(0)}$. Thus, accurate segmentation can be obtained using the label decoder. To test the robustness of LDSeg to noise, we have generated the noisy image $I_{\sigma}$ from the input $I$ by, 
\begin{equation}
    I_{\sigma} = I + \mathcal{N}(0,\sigma)
\end{equation}
where $I$ is a sample from test data and $\sigma$ is the noise variance. \ref{Figure8}(a) shows the DSC scores for LDSeg and a deterministic model ResUNet against different variances of added noise on the Echo and the Knee datasets, respectively. LDSeg showed strong robustness to the added noise even for $\sigma=0.2$, and maintained reasonably good segmentation accuracy throughout. In contrast, the accuracy for ResUNet dropped drastically with the increasing amount of noise added to the source image. Figure \ref{Figure8}(b) shows a sample Echo image with added noise of different variances and the corresponding segmentation results by LDSeg and ResUNet.

\section{Ablation Study}
\begin{figure}[ht]
    \centering
    \includegraphics[width=0.48\textwidth]{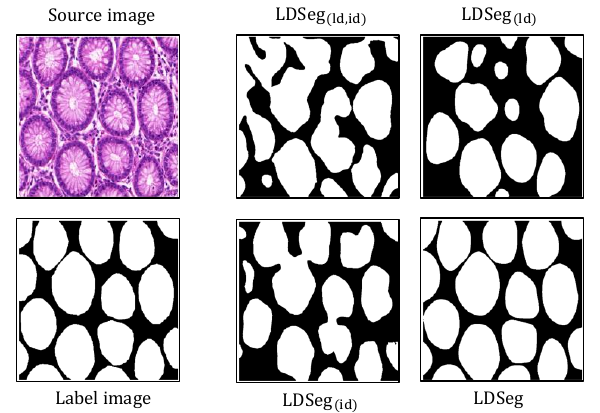}
    \caption{A sample test image along with its ground truth mask and predictions by different variants of LDSeg.}
    \label{Figure9}
\end{figure}

Two major components that distinguish the proposed LDSeg from other diffusion-based segmentation models are the label and the image encoders, which learn the latent embeddings of the object shape manifolds and the source image, respectively. We tested the effectiveness of each of these two components by creating several variants of LDSeg:
\begin{itemize}
    \item LDSeg: The proposed framework that uses both the label and the image encoder.
    \item LDSeg\textsubscript{(ld)}: The label encoder is replaced with a label down-sampler that down-samples the label image to the same size of $z_{l(0)}$. The image encoder is unchanged.
    \item LDSeg\textsubscript{(id)}: The image encoder is replaced with an image down-sampler that down-samples source image to the same size of $z_i$. The label encoder is unchanged.
    \item LDSeg\textsubscript{(ld,id)}: Both the label and the image encoder are replaced with the label and the image down-samplers.
\end{itemize}
Table \ref{table-5} shows the results of the ablation study in the GlaS data set. The models with direct down-sampling by nearest-neighbor interpolation of the image or/and label performed poorly. This indicates that the denoiser trained on low-dimensional latent space may have superior noise prediction capability in general, but without proper learning of the object shape manifolds along with the robust imaging features, generating accurate segmentation is extremely challenging. Figure \ref{Figure9} shows example segmentations with different variants of the LDSeg models.
\section{Discussion}
In medical imaging, often times an image consists of 3D scans and cannot be down-sampled without loosing important imaging features due to the complex tissue structures, organ-to-organ  interactions, etc. LDSeg can be directly applied on  large 3D images for accurate segmentation, while other traditional DPM may not be even implementable due to the lack of enough GPU/CPU memory. On top of that, fast sampling in the reverse process makes the proposed LDSeg computationally much efficient. This can be attributed to the much simpler sampling objective of LDSeg compared to a traditional DPM used for image generation. LDSeg samples from posterior distribution for segmentation generation, which is much simpler and concentrated than a prior distribution in image generation case, resulting in remarkably faster sampling. Moreover, the end-to-end training strategy enables robust segmentation related representation learning in the latent space, further improving the sampling efficiency. 

\begin{figure}[ht]
    \centering
    \includegraphics[width=0.48\textwidth]{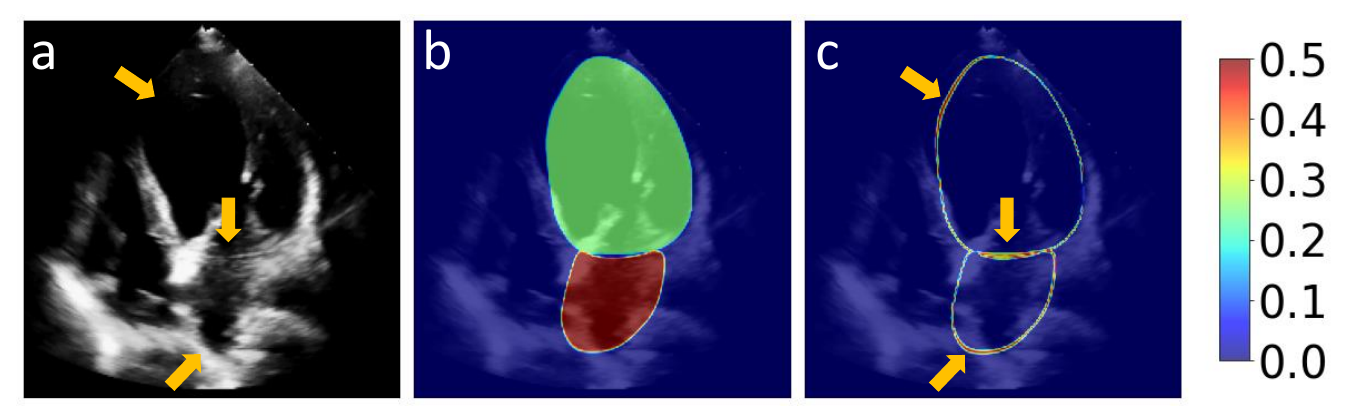}
    \caption{An example of uncertainty estimation of segmentation in the Echo dataset. \textbf{(a)} A sample Echo frame with marked unclear LV and LA wall regions (orange arrows). \textbf{(b)} The mean segmentation map using 100 sampling runs. \textbf{(c)} The obtained standard deviation (SD) map from the 100 sampling runs. The orange arrows show the highly uncertain regions with three maximum SDs that correlate to the locations in (a)}
    \label{Figure10}
\end{figure}

LDSeg is significantly more robust to noises presented in the source images than the traditional deterministic segmentation models, which mitigates the noisy image acquisition problem in medical imaging. A key challenge for the deterministic segmentation models is to measure prediction uncertainty. Being generative in nature, LDSeg can estimate prediction uncertainty by obtaining standard deviation of the predictions with multiple runs. Figure \ref{Figure10} shows an example of uncertain regions on object boundary estimation using LDSeg.

A possible limitation of the proposed approach is the use of a low-dimensional image embedding learned for highly complex medical structures. As the data complexity increases in terms of tissue structures with various distributions, it is very challenging to learn a proper image embedding preserving all the fine details, which may hamper the denoising process of the denoiser. One way to address this problem would be to learn different frequency patterns of the input images by the image encoder to enforce additional conditions on the denoiser.

\section{Conclusion}
Adapting DPMs for medical image segmentation poses significant challenges due to large image sizes, complex tissue structures, and noisy image acquisitions. We present LDSeg, a novel latent diffusion based segmentation framework that leverages the learned low-dimensional latent representations of the image and target object's shape manifolds in an end-to-end training strategy, which substantially improves the training and inference efficiency for not only 2D, but also 3D and higher-dimensional medical image segmentation. The proposed LDSeg demonstrated much improved robustness to severe noises presented in the source image.

\bibliographystyle{IEEEtran}
\bibliography{tmi}

\end{document}